%% file: main.tex
\documentclass{article} %
\usepackage{iclr2027_conference,times}

\input{math_commands.tex}

\usepackage{hyperref}
\usepackage{url}
\usepackage{placeins}

\input{preamble}

\title{Co${}^{2}$Skill: Whole-Body Control via Skill Composition for Long-Horizon Human-Environment Interaction}

\author{\parbox[t]{\dimexpr\textwidth-2\tabcolsep\relax}{\centering\normalfont
Jeonghwan Kim\qquad Hyeonwoo Kim\qquad Hanbyul Joo\\[0.6em]
Department of Computer Science\\
Seoul National University
}}

\iclrfinalcopy %
\begin{document}

\maketitle
\lhead{}                          %
\renewcommand{\headrulewidth}{0pt} %
\suppressfloats[t]

\begin{abstract}
Achieving human-level dexterity in complex, unstructured environments requires the seamless integration of whole-body scene interaction and dexterous object manipulation skills. 
While existing physics-based controllers generate physically plausible behaviors in each domain, they largely address these two capabilities independently. 
In this paper, we present Co${}^{2}$Skill that integrates scene interaction and dexterous manipulation through a unified policy formulation. 
Built on a pretrained motion prior, the policy uses task and phase dependent observation masks to select information relevant to the current interaction goals.
We introduce a goal-conditioned loco-manipulation curriculum that combines partial reference guidance for precision with exploration from varied initial states while allowing goal-directed execution beyond the demonstrated trajectories. 
We further introduce a cross-task curriculum that jointly trains individual skills and selected task sequences, preserving physical states across task boundaries and maintaining grasps during subsequent scene interactions. 
Together, these support sequential task execution and simultaneous scene interaction with object manipulation. 
We evaluate sitting, standing, climbing, stair traversal, and goal-directed manipulation, together with sequential execution and with random different conditions. Additionally, we demonstrate skill compositions in indoor environments, illustrating their integration within the same control formulation.

\end{abstract}

\input{sec/01_introduction}

\input{sec/02_related}
\input{sec/03_prelim}

\input{sec/04_method}

\input{sec/05_experiments}
\input{sec/06_results}
\input{sec/07_conclusion}

\bibliography{iclr2027_conference}
\bibliographystyle{iclr2027_conference}

\end{document}

%% file: math_commands.tex
\usepackage{amsmath,amsfonts,bm}

\def\eqref#1{equation~\ref{#1}}

\def\1{\bm{1}}

\DeclareMathAlphabet{\mathsfit}{\encodingdefault}{\sfdefault}{m}{sl}
\SetMathAlphabet{\mathsfit}{bold}{\encodingdefault}{\sfdefault}{bx}{n}

%% file: preamble.tex
\usepackage{xcolor}
\usepackage{booktabs}
\usepackage{multirow}

\usepackage{graphicx}   %

\providecommand{\Description}[1]{}

\usepackage{xspace}
\newcommand{\eg}{e.g.,\xspace}

%% file: sec/01_introduction.tex
\section{Introduction}
\label{sec:intro}

Everyday activities at home rarely consist of a single skill. A person may walk to a shelf, pick up a phone, and then sit down on a chair while still holding it. For a simulated humanoid, this means coordinating locomotion, whole-body interaction with the surrounding scene, and finger-level grasping across a room-scale space rather than on an empty floor or a tabletop. Each skill must also remain physically stable when it starts from whatever state the previous skill leaves behind.

Physics-based character control with deep reinforcement learning has made steady progress toward such behavior. Tracking policies trained on large motion capture collections now imitate a wide range of human movement~\citep{peng2018DeepMimic, luo2023Perpetual, tessler2024MaskedMimic}, and motion priors distilled from these policies can be reused for new downstream tasks~\citep{luo2024Universal}. Building on such learned motor skills, two lines of work have developed largely in isolation. Human-scene interaction (HSI) controllers move the whole body relative to furniture and terrain, for example to sit on a chair or climb a staircase~\citep{xiao2024Unified, pan2025TokenHSI}, while the hands (with no fingers attached) play little role. Human-object interaction (HOI) controllers achieve precise finger contact and object control~\citep{luo2025Omnigrasp, tessler2025MaskedManipulator}, but they typically operate close to the object and often depend on dense reference trajectories from datasets such as GRAB~\citep{taheri2020GRAB}. 
Ideally, a single controller should both interact with the surrounding scene and manipulate the objects within it.

Combining these capabilities is harder than simply putting the two kinds of skills together. First, scene interaction and manipulation depend on different inputs. Scene interaction needs scene geometry and spatial targets, while manipulation needs object motion, hand poses, and contact. Second, each skill usually learns from a fixed set of starting states, which differ from the states it meets in a longer task. A sitting policy that learns from a normal standing pose never starts with an object in its hands, and a grasping policy mostly starts from the pose in its reference clip. Switching between separate specialist policies does not solve this problem, and it cannot handle cases where a scene task and a grasp must be performed at the same time. Third, reference motions give the detailed guidance needed for precise contact or large body adjustment, but they cover only the situations present in the motion capture data. A controller that relies on them therefore struggles to approach a chair or an object from new positions.

We address these challenges with our Co${}^{2}$Skill, a physics-based controller that performs scene interaction and dexterous manipulation with a single policy. We build the policy on top of a pretrained motion prior and decoder~\citep{luo2024Universal}. To handle the different inputs of each task, we design a transformer that reads task tokens for the character, scene, object, and hands, enabling only those relevant to the current task and phase. To handle the gap between training and initial states and the limited coverage of reference motions, we introduce two curricula. The goal-conditioned loco-manipulation curriculum uses reference motions to guide close-range interactions such as sitting down or forming a grasp, while exploration from varied starting states teaches the policy to reach the same goals beyond the reference. The cross-task curriculum trains individual tasks together with task sequences, such as lifting an object and then sitting while holding it, so that the policy learns from the states that sequential execution actually produces. 
Our shared policy outperforms MaskedMimic across five scene-interaction tasks and achieves success rates matching or exceeding per-task specialists on four of five tasks. We also evaluate pick-move-place task and qualitatively demonstrate manipulation-to-scene transitions while maintaining object grasps.

\begin{figure}[t]
  \centering
  \includegraphics[width=\textwidth]{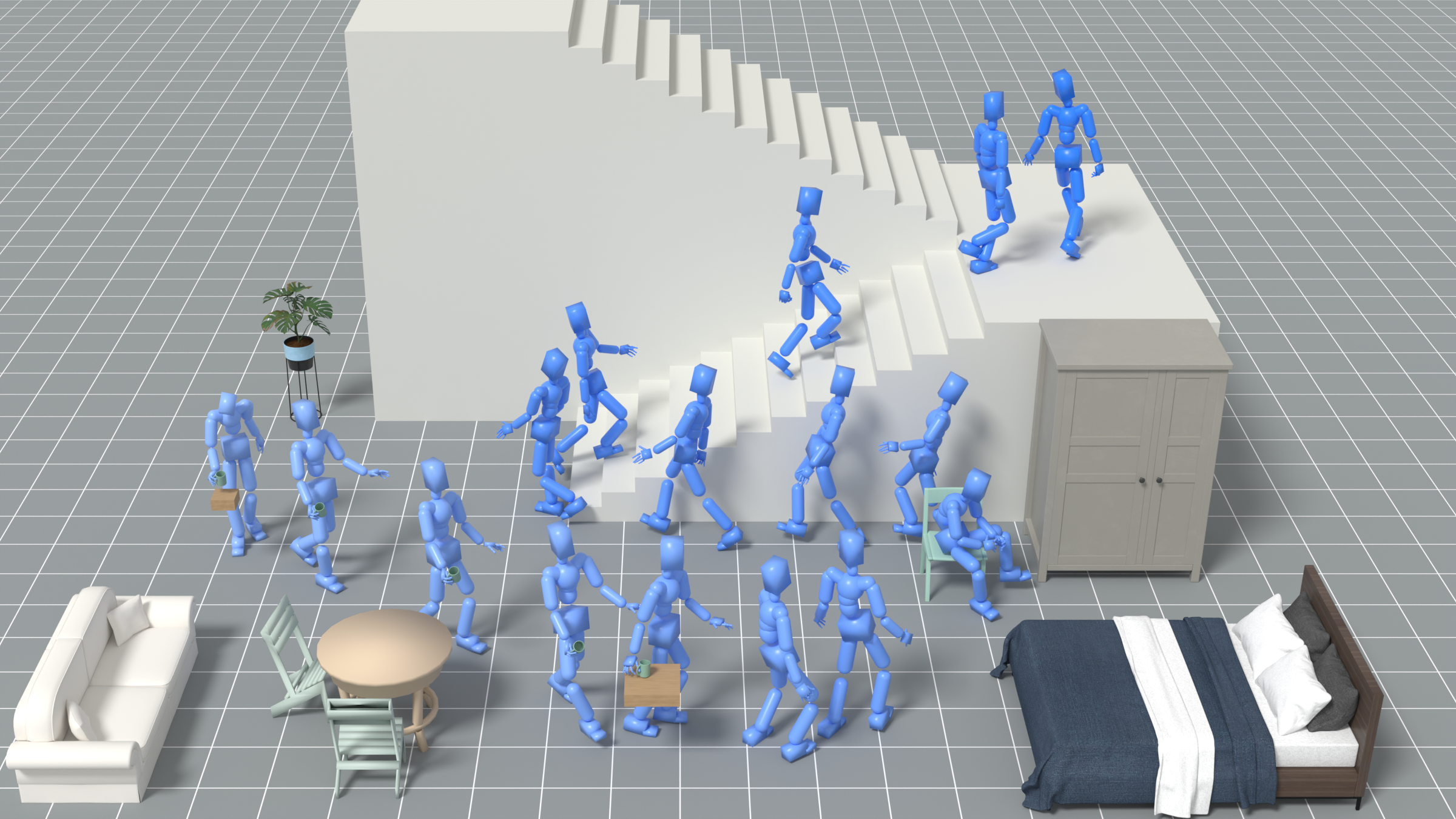}
  \caption{Our method effectively deals with both scene interactions and dexterous manipulations.}
  \Description{Sequence visualizations of a simulated character performing locomotion and dexterous manipulation tasks across diverse 3D scenes, all controlled by a single unified policy.}
  \label{fig:teaser}
\end{figure}

Our contributions are summarized as follows:
\begin{itemize}
    \item \textbf{A Unified Framework for Scene Interaction and Dexterous Manipulation.} We introduce a unified control framework that combines whole-body scene interaction and dexterous object manipulation within a single goal-conditioned policy.
    \item \textbf{A Goal-Conditioned Loco-Manipulation Curriculum.} We combine reference-guided interaction learning with exploration from varied initial states, using partial motion guidance to learn precise interactions while allowing goal-directed execution beyond the reference trajectory.
    \item \textbf{A Cross-Task Curriculum for Sequential and Simultaneous Interactions.} We jointly train individual tasks and selected task sequences, preserving physical states across transitions and maintaining grasp conditions during subsequent scene interactions.
\end{itemize}

%% file: sec/02_related.tex
\section{Related Work}
\label{sec:rel}

\subsection{Human-Scene Interactions}
Understanding human-scene interactions requires datasets that faithfully capture spatial relations between human motion and environmental context. Early mocap datasets of humans interacting with static environments~\citep{starke2019Neural, hassan2021Stochastic} enabled training models to predict behavior given goal targets. Interaction Replica~\citep{guzov2022Interaction} extended this by tracking interactions with partially dynamic scenes. Locomotion-Action-Manipulation~\mbox{\citep{lee2023LocomotionActionManipulation}} synthesizes human-scene interactions in 3D environments by coordinating locomotion with manipulation actions. More recent work augments datasets by capturing human behavior and applying VR rendering to diversify indoor environments~\citep{jiang2024Scaling, jiang2024Autonomous}. These datasets establish foundations for learning human behavior in environments.

\subsection{Human-Object Interactions}
Human behavior naturally incorporates manipulation of objects in the surrounding environment. BEHAVE~\citep{bhatnagar2022BEHAVE} and Object Motion Guided Human Motion Synthesis~\citep{li2023Object} capture human body motion with dynamic objects (\eg, suitcases) without hand grasping. Hand-only datasets~\citep{chao2021DexYCB, zhang2021ManipNet} focus on finger grasping sequences without body motions. GRAB~\citep{taheri2020GRAB} and ARCTIC~\citep{fan2023ARCTIC} capture whole-body manipulation with hand articulation but limited action diversity. ParaHome~\citep{kim2025ParaHome} captures indoor sequences with both locomotion and finger-level manipulations. For kinematic synthesis, IMoS~\citep{ghosh2023IMos} generates intent-driven full-body motions for object interactions, and FORCE~\citep{zhang2024FORCE} introduces physics-aware synthesis with force modeling.

\subsection{Physics-Based Character Control}
DeepMimic~\citep{peng2018DeepMimic} demonstrated that deep RL can successfully track reference motion in physics simulation. Adversarial reward settings~\citep{peng2021AMP, xu2021GANLike, zhang2025PhysicsBased} introduced stylized control without explicit reference tracking, while latent-based methods~\citep{peng2022ASE, dou2023C$cdot$ASE, tessler2023CALM} learn reusable skill-embeddings for compositional task execution. Recent work scaled motion tracking with large-scale datasets~\citep{AMASS:ICCV:2019} to learn single policies for real-time tracking~\citep{luo2023Perpetual}, with adaptations for language control~\citep{juravsky2022PADL, juravsky2024SuperPADL} and multi-task composition via masked observations~\citep{tessler2024MaskedMimic}. For physics-based manipulation, D-Grasp ~\citep{christen2022DGrasp} synthesizes dynamic grasps for hand-object interactions, and PhysHOI~\citep{wang2023PhysHOI} tackles physics-based imitation of dynamic human-object interaction. For scene-level interactions, InterPhys~\citep{hassan2023Synthesizing} synthesizes physical character-scene interactions with contact handling, UniHSI~\citep{xiao2024Unified} performs HSI using chain-of-contacts with LLM planning, and TokenHSI ~\citep{pan2025TokenHSI} demonstrates multi-task composition with task tokenization.

\subsection{Behavior Foundation Models}
A behavior foundation model~\citep{tirinzoni2025ZeroShot} refers to a universal policy that generalizes across diverse motion styles trained with large-scale datasets.
MoConVQ~\citep{yao2023MoConVQ} uses discrete representations as compact motion priors that interface with language models. PULSE~\citep{luo2024Universal} trains a universal policy with CVAE aimed for adaptation to diverse downstream tasks. More recent work utilizes task-specific tokens as flexible representations to adapt general tracking policies~\citep{tessler2024MaskedMimic, vainshtein2025Task}. Composite motion learning~\citep{xu2023Composite} enables task-aware skill composition for complex behaviors. Based on these advances, we aim to integrate locomotion datasets~\citep{AMASS:ICCV:2019} and whole-body manipulation datasets~\citep{taheri2020GRAB} through a staged training strategy that learns a general motion prior and then steer the trained model to perform various tasks.

%% file: sec/03_prelim.tex
\section{Preliminaries}
\label{sec:prelim}

Given a 3D scene with static structures such as furniture and movable objects, our goal is to 
output a joint-level action target
that interacts with the scene and manipulates objects within it. The resulting motion coordinates locomotion, scene contact, and hand-object interaction to reach task goals and to move between selected skills, without a large-scale reference motion dataset that covers the entire demonstration sequence.

\subsection{Formulation}

We formulate character control as a goal-conditioned reinforcement learning problem. Let $s_t \in \mathcal{S}$ denote the simulation state, including the character and its environment, $g_t \in \mathcal{G}$ the task condition, and $a_t \in \mathcal{A}$ the joint-level control action. The task condition specifies the active interaction and its targets, and may change as the state progresses. 
Physical transitions follow $P(s_{t+1}\mid s_t,a_t)$, while task conditions are updated according to each task's progression rules.

At each control step, the policy selects an action $a_t \sim \pi(\cdot\mid s_t,g_t)$ and receives a task-dependent reward $r_t=R(s_t,a_t,g_t)$. Our objective is to learn a policy that maximizes the expected discounted return over an episode of $T$ control steps:
\begin{equation}
    \pi^* = \arg\max_{\pi} \mathbb{E}_{\pi} \left[ \sum_{t=0}^{T-1} \gamma^t R(s_t, a_t, g_t) \right],
\end{equation}
where $\gamma \in [0,1)$ is the discount factor and the expectation is over task and initial-state sampling and the resulting trajectories.

We use an SMPL-X-based humanoid with 52 joints~\citep{SMPLX}, including a floating root, following~\citep{luo2023Perpetual}. Each of the 51 non-root joints has three actuated degrees of freedom, giving a joint-level action space $\mathcal{A}\subset\mathbb{R}^{153}$. The action $a_t$ specifies joint position targets tracked by proportional--derivative (PD) controllers; the root's global translation and rotation evolve through the simulated dynamics. Here, $\pi$ denotes the complete controller producing these joint-level actions; its motion-prior-based parameterization is given in Eq.~\ref{eq:latent-control}.

%% file: sec/04_method.tex
\section{Method}
\label{sec:method}

\begin{figure}[t]
  \centering
  \includegraphics[width=\textwidth]{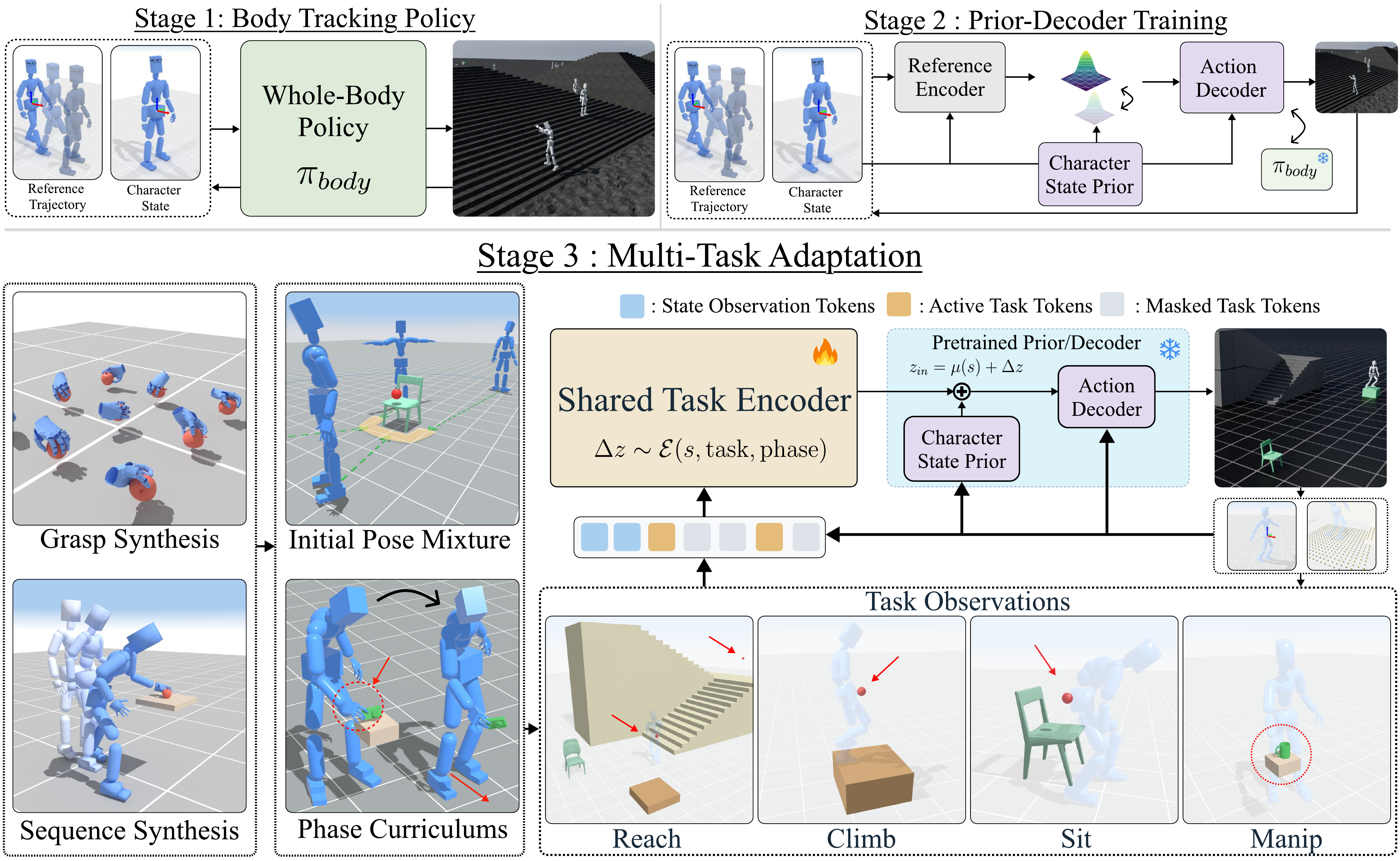}
  \caption{\textbf{Overview.} We first learn a motion prior and decoder from a whole-body tracking policy, then adapt a shared task encoder for scene interaction and manipulation tasks.}
  \label{fig:overview}
\end{figure}

Our framework (Fig.~\ref{fig:overview}) brings scene interaction and dexterous manipulation into a shared control formulation, with task conditions specifying spatial goals and phase-dependent interaction requirements. 
For scene interaction, these conditions specify movement or contact goals, such as a pelvis target on a chair for sitting.
For manipulation, they specify the desired object pose, active hands, object-relative grasp targets, and whether each hand should maintain or release its grasp. 
To leverage a learned whole-body motion prior for scene interaction and manipulation, we build on the pretrained prior and decoder of PULSE~\citep{luo2024Universal}.
Given proprioceptive observations and the previous action, denoted together by $x_t$, the prior predicts a distribution over latent codes that the decoder maps to joint-level action targets. Its mean, $\mu_{\theta}(x_t)$, provides the state-conditioned base code.
A transformer uses character, environment, and task observations to predict a Gaussian latent residual $\Delta z_t$, which adjusts the prior mean before decoding:
\begin{equation}
    z_t = \mu_{\theta}(x_t) + \Delta z_t,
    \qquad a_t = D_{\theta}(x_t,z_t).
    \label{eq:latent-control}
\end{equation}
The prior and decoder remain fixed while the task-conditioned policy is trained.

\subsection{Unified Interaction Task Representation}
\label{subsec:interaction-representation}

\paragraph{Task tokens.}

A shared policy must accommodate different interaction conditions and combine them when scene interaction and manipulation are required simultaneously. 
We therefore organize observations and task conditions into groups according to their roles and map each group $o_i$ to a fixed-dimensional token $h_i=T_i(o_i)$ using its corresponding tokenizer $T_i$, following TokenHSI~\citep{pan2025TokenHSI}. Shared tokens encode the character's proprioceptive state, previous action, motion-prior mean, and local scene heightmap. Scene-interaction tokens encode spatial goals and the poses and bounding boxes of interacting objects. Manipulation tokens encode the manipulated object's state, geometry, and target pose. Per-hand tokens additionally encode the desired grasp as an object-relative wrist pose and the corresponding wrist target expressed relative to the character.

\paragraph{Masked task conditioning.}
Each token $h_i$ is associated with a binary mask $m_i$, where $m_i=0$ excludes the token from attention. Shared tokens remain active, while task tokens are enabled according to the current task. For example, sitting while holding an object activates scene-interaction tokens together with object and grasp tokens. The transformer defines a distribution over latent residuals conditioned on the token sequence and its corresponding masks:
\begin{equation}
    \Delta z_t
    \sim \mathcal{E}\!\left([h_1,\ldots,h_N],\,[m_1,\ldots,m_N]\right).
    \label{eq:task-conditioning}
\end{equation}

\subsection{Goal-Conditioned Loco-Manipulation Curriculum}
\label{sec:loco_manipulation_curriculum}
The task representation specifies the desired interaction, but learning to map it to the actual action requires known coordinated body motion and precise contact formation. Reference motions provide a dense guidance for these behaviors, but
do not specify how to achieve the same goal from different initial states or object placements. We therefore combined reference-guided and goal-driven learning, varying the scope of reference guidance according to the task. 

\paragraph{Goal-conditioned scene interaction.}
We formulate standing up from a chair, descending from a box, and traversing stairs as Reach tasks, using intermediate waypoints as reward where needed. For Sit and Climb-up, we jointly train the same policy in reference and exploration environments. Reference environments are initialized relative to the demonstration and provide tracking supervision for precise interaction. 
Exploration environments instead sample starting positions at varying distances and directions around the furniture, requiring the character to approach and complete the same interaction from states beyond the demonstrated approach. 

\paragraph{Pick-move-place manipulation.}
We train pick-move-place using an end-target object pose $G_{o,t}^{*}$ and grasp/release conditions, rather than the object trajectories used by OmniGrasp~\citep{luo2025Omnigrasp}.
The object goal alone does not determine how the hands should grasp the object or when they should disengage. We therefore provide an object-relative grasp wrist pose $G_{w\mid o}^{h,*}$ for each active hand $h\in\{L,R\}$. Given the current object pose $G_{o,t}$, the corresponding world-space wrist target is
\begin{equation}
    G_{w,t}^{h,*}=G_{o,t}G_{w\mid o}^{h,*}.
    \label{eq:grasp-wrist-target}
\end{equation}
This object-relative specification preserves the desired hand and object alignment across object placements, and the resulting wrist targets are expressed relative to the character for policy input. Per-hand binary conditions $b_t^h\in\{0,1\}$ indicate whether each active hand should retain or release its grasp. These targets and per-hand conditions form the manipulation component of $g_t$ and are updated as execution progresses through grasp formation, lifting, transport, placement, and release. 
During placement, we sample a dwell interval in which the hands retain their grasp before the release condition is activated, distinguishing arrival at the target pose from completion of placement and release.
Reference motion guides initial body coordination and wrist to finger alignment during the close range grasp formation. At the reference lift time, we remove full-body tracking and continue with task-specific goals, while actual lift completion determines when transport begins.

\subsection{Cross-Task Curriculum}
\label{sec:cross_task_curriculum}

\paragraph{Joint training of individual and sequential tasks.}
We jointly train individual scene and manipulation tasks with selected manipulation-to-scene sequences, in which grasp-and-lift is followed by sitting, climbing, or reaching. Once lifting is complete, we activate the destination scene goal and its task tokens without resetting the character or object state. The subsequent task therefore begins from the state produced by lifting, rather than an independently sampled initialization. Without a reference motion spanning the sequence, the policy explores how to continue into the scene task under task rewards.

\paragraph{Maintaining an interaction across tasks.}
The scene task must be performed while preserving the grasp established during lifting. We therefore keep object and grasp tokens enabled alongside the scene-task tokens, with grasp-retention conditions $b_t^h=1$ for the active hands. We define the carried object's target $G_{o,t}^{*}$ from its pose relative to the character's heading frame at handover and update it with the character's motion. Placement and release remain inactive during this segment. 
These conditions jointly guide scene interaction and grasp retention.

%% file: sec/05_experiments.tex
\section{Experiments}
\label{sec:exp}

We evaluate individual scene-interaction and manipulation capabilities, followed by their sequential and simultaneous execution. For scene interaction, we compare our shared policy with per-task specialists and an adapted MaskedMimic baseline. We also assess the effect of synthetic data augmentation through reference-conditioned lifting with MimicManipulator, and evaluate our policy on goal-conditioned pick-move-place tasks without prescribed object trajectories. Finally, we qualitatively demonstrate manipulation-to-scene transitions and scene interaction while maintaining an object grasp.

\subsection{Setups}
\label{subsec:setups}

\paragraph{Datasets.}
\label{subsec:tasks-data}

We use AMASS~\citep{AMASS:ICCV:2019} and GRAB~\citep{taheri2020GRAB} dataset throughout the tasks.
To enrich finger articulation, we augment AMASS motions with grasping segments sampled from GRAB.
For HOI training, we combine mocap references, including their object poses, with synthetic sequences generated using Dexonomy~\citep{chen2025dexonomy} and Kimodo~\citep{rempe2026kimodo}.
We filter the synthetic sequences to remove unnatural whole-body motions during object manipulation. 
For reference-conditioned lifting and holding, we use grasp-and-lift segments from GRAB.
For scene interactions, we use motion data, human-object interaction intervals, and object geometry provided by TokenHSI~\citep{pan2025TokenHSI}.

\paragraph{Simulation setups.}
We use the ProtoMotions~\citep{tessler2024MaskedMimic} framework with
IsaacLab~\citep{mittal2025isaaclab}. Policies operate at 30\,Hz, with physics frequency
at 120\,Hz.
During tracker training and prior distillation, we apply random pushes with randomized wrist
and finger masses to improve robustness for downstream object manipulations as well as complex terrains with pyramidal-stairs and hills for the ease of the scene interaction.

\paragraph{Policy architecture.}
Following PULSE~\citep{luo2024Universal}, we use MLPs for the motion prior
and decoder. Both remain frozen during task-policy training.
The task encoder is a four-layer transformer with four attention heads and
512-dimensional tokens.

\paragraph{Baselines.}
For scene interaction, we adapt MaskedMimic~\citep{tessler2024MaskedMimic} to our task environments by adding observations of surrounding objects and distilling a tracking teacher policy. We also train a separate instance of our task policy for each scene task, denoted as per-task specialists to assess the quality of sharing a policy across tasks. For manipulation, we implement MimicManipulator~\citep{tessler2025MaskedManipulator} which is a reference trajectory conditioned tracker to evaluate synthetic data augmentation comparing training on mocap references along with training on mocap and synthetic references. We additionally use its reference-conditioned lifting performance to gauge the picking performance of our goal-conditioned pick-move-place policy.

\subsection{Metrics}
\label{subsec:setup}

For scene-interaction tasks, we report task success rate (SR, \%) and
target position error (cm). All distances below are measured between
the interacting targets and the goal position.

\paragraph{Success rate.}
\emph{Sit.} Each chair has a predefined sitting target. Success requires
the pelvis to remain within 10\,cm of this target for 60 consecutive
control steps. 
\emph{Climb-up.} Success is achieved when the pelvis is within 30\,cm
of the target above the box center.
\emph{Stand-up.} The character starts seated on a chair. Success requires
the pelvis to be at least 85\,cm above the ground and simultaneously
within 30\,cm of the goal before timeout.
\emph{Climb-down.} The character starts on top of a box. Success requires
the pelvis to reach within 30\,cm of the ground-level goal without falling.
Neither Stand-up nor Climb-down requires a holding interval at the end.
\emph{Stair.} The character starts on the ground near the staircase.
Success requires first reaching within 30\,cm of the middle waypoint,
then remaining within 30\,cm of the final waypoint. 
The reported Stair SR measures completion of this full sequence.

\paragraph{Target-position error.} For Sit and Stair tasks, we average the error
over the first successful holding interval, using the seat target and
the final goal point. For Climb-up, Stand-up, and Climb-down tasks,
we measure the error at the first successful control step. We then
average these per-episode errors over successful episodes with equal
weights.

\paragraph{HOI reference lifting.}
We evaluate reference-conditioned lifting with success rate, mean per-joint position error (MPJPE), and tracking length. Each reference is truncated at the first frame where the object rises at least 10\,cm above its initial height. A trial succeeds if tracking reaches this endpoint without any humanoid rigid-body position error exceeding 0.5\,m or object position error exceeding 0.25\,m, and without numerical failure. MPJPE measures world-space position error over all humanoid rigid bodies, averaged over valid tracking frames within each trial and then across trials; terminal failure frames are excluded. Tracking length measures elapsed control time until the first tracking failure or reference completion, including the terminal detection step, averaged across trials.

\paragraph{Pick-move-place.}
We report pick, move, and release success rates at each phase. Pick and move require hand object contact and object position errors below 3\,cm and 5\,cm. Release is commanded after a 2\,s dwell in the placement target region and succeeds after 15 contact-free steps within 5\,cm of the target, using a 1\,N contact threshold. Contact loss for 20 consecutive steps before the command or continued contact for 15 steps afterward causes failure. Release error is averaged over the successful 15-step window and across successful episodes.

%% file: sec/06_results.tex
\subsection{Quantitative Results}
\label{sec:results}

\begin{table}[!t]
  \centering
  \scriptsize
  \setlength{\tabcolsep}{2pt}
  \renewcommand{\arraystretch}{1.15}
  \begin{tabular*}{\linewidth}{@{\extracolsep{\fill}}l*{10}{c}@{}}
    \toprule
    \multicolumn{1}{c}{\multirow{2}{*}{Method}}
      & \multicolumn{2}{c}{Sit}
      & \multicolumn{2}{c}{Climb-up}
      & \multicolumn{2}{c}{Reach(Stand-up)}
      & \multicolumn{2}{c}{Reach(Climb-down)}
      & \multicolumn{2}{c}{Reach(Stair)} \\
    \cmidrule(lr){2-3}\cmidrule(lr){4-5}\cmidrule(lr){6-7}\cmidrule(lr){8-9}\cmidrule(l){10-11}
      & SR (\%) & Error [cm] & SR (\%) & Error [cm]
      & SR (\%) & Error [cm] & SR (\%) & Error [cm]
      & SR (\%) & Error [cm] \\
    \midrule
    MaskedMimic & 44.27 & 8.36 & 93.23 & 28.72 & 72.40 & 29.55 & 83.33 & 29.50 & 7.81 & 29.07 \\
    \midrule
    Ours (Per-Task Specialists) & \textbf{99.48} & \textbf{5.82} & \textbf{100.00} & 27.74 & 98.31 & \textbf{23.93} & \textbf{100.00} & \textbf{26.79} & 54.41 & 17.59 \\
    Ours & 93.22 & 6.68 & \textbf{100.00} & \textbf{27.73} & \textbf{100.00} & 27.62 & \textbf{100.00} & 27.38 & \textbf{69.23} & \textbf{11.58} \\
    \bottomrule
  \end{tabular*}
  \caption{\textbf{Scene interaction tasks.} For each task, our per-task policy outperforms the baseline while our task-integrated policy works on par with each expert policy at each task.}
  \label{tab:single-skill}

  \vspace{1em}
  \centering
  \scriptsize
  \setlength{\tabcolsep}{2pt}
  \renewcommand{\arraystretch}{1.15}
  \begin{minipage}[t]{0.52\linewidth}
    \centering
    \textbf{(a) HOI Lifting Task} \\[2pt]
    \begin{tabular*}{\linewidth}{@{\extracolsep{\fill}}llccc@{}}
      \toprule
      Split & Training data & \shortstack{\strut SR\\\strut (\%) $\uparrow$}
        & \shortstack{\strut MPJPE\\\strut {}[cm] $\downarrow$}
        & \shortstack{\strut Length\\\strut {}[s] $\uparrow$} \\
      \midrule
      Train & GRAB Only & \textbf{93.66} & \textbf{2.42} & \textbf{2.26} \\
            & GRAB + Synthetic & 83.41 & 3.04 & 2.25 \\
      \midrule
      Test  & GRAB Only & 56.67 & 3.29 & 1.78 \\
            & GRAB + Synthetic & \textbf{90.00} & \textbf{3.11} & \textbf{1.99} \\
      \bottomrule
    \end{tabular*}
  \end{minipage}\hfill
  \begin{minipage}[t]{0.46\linewidth}
    \centering
    \textbf{(b) Pick-Move-Place Task}\\[2pt]
    \begin{tabular*}{\linewidth}{@{\extracolsep{\fill}}lcccc@{}}
      \toprule
      Split & \shortstack{\strut Pick SR\\\strut (\%) $\uparrow$}
        & \shortstack{\strut Move SR\\\strut (\%) $\uparrow$}
        & \shortstack{\strut Release SR\\\strut (\%) $\uparrow$}
        & \shortstack{\strut Release Error\\\strut {}[cm] $\downarrow$} \\
      \midrule
      \addlinespace[2.837pt]
      Train & 81.02 & 77.29 & 63.39 & 2.34 \\
      
      \addlinespace[5.673pt]
      Test & 84.80 & 75.20 & 64.80 & 2.49 \\
      
      \addlinespace[5.673pt]
      Unseen & 47.90 & 40.25 & 25.19 & 2.87 \\
      \addlinespace[2.837pt]
      \bottomrule
    \end{tabular*}
  \end{minipage}
  \caption{\textbf{Reference tracking and pick-move-place evaluation.}
    (a) Effect of synthetic training data augmented to mocap data and trained on the tracking policy.
    (b) Picking, moving, and releasing the given object, we measure success rates at each conditioned interact phase and measure release error of the object at the goal pose.}
  \label{tab:hoi-reference-lifting}
\end{table}

\paragraph{Scene-Interaction Task Comparison.}
\label{subsec:task-performance}
\label{subsec:hoi-reference-lifting}

Table~\ref{tab:single-skill} compares our shared policy and per-task specialists with MaskedMimic. Both variants achieve higher success rates and lower target-position errors than MaskedMimic across all five tasks. The shared policy achieves 100.00\% SR on Climb-up, Stand-up, and Climb-down, and 93.22\% on Sit. Stair remains the most challenging task, with an SR of 69.23\%, compared with 7.81\% for MaskedMimic.
Compared with per-task specialists, the shared policy improves Stair SR from 54.41\% to 69.23\% and reduces error among successful episodes from 17.59 to 11.58\,cm. It also increases Stand-up SR from 98.31\% to 100.00\% and matches specialist SRs on Climb-up and Climb-down, while Sit SR decreases from 99.48\% to 93.22\%. These results show that a shared policy supports multiple scene interactions with task-dependent trade-offs.

\paragraph{Data Augmentation Justification.}

In Table~\ref{tab:hoi-reference-lifting}(a), we train MimicManipulator on the GRAB dataset and its augmented version. On train dataset, the model trained with only GRAB dataset show higher SR with lower MPJPE compared to the reference used. However on test set, the result differs by large margin compared to train set result, substantially improving the SR from 56.67\% to
90.00\%. MPJPE on test set also reduces from 3.29 to 3.11\,cm and the average tracking length increases from 1.78 to 1.99\,s. 
Such train-test gap was also observed in MimicManipulator~\citep{tessler2025MaskedManipulator}, and we believe the data augmentation can ease such discrepancy.
Overall, the result suggests that the additional diversity introduced by the synthetic sequences reduces overfitting to the training references and improves generalization to unseen motions with known object identities.

\paragraph{Pick-Move-Place Evaluation.}
Table~\ref{tab:hoi-reference-lifting}(b) reports cumulative pick, move, and place success rates, each normalized by the total number of evaluation episodes. Our policy achieves a training-set Pick SR of 81.02\%, compared with 83.41\% SR for the reference-conditioned tracker trained with augmented data. This comparison provides context rather than a direct comparison, since the tracker measures reference completion whereas pick requires maintained contact and proximity to the lift target, as well as difference in conditioning of trajectory.

The policy completes the full pick-move-place sequence in 63.39\% and 64.80\% of train and test episodes. On unseen objects, pick, move, and release success rates decrease to 47.90\%, 40.25\%, and 25.19\%, showing partial transfer with a substantial drop in overall completion. Among successful release episodes, mean placement errors remain between 2.34 and 2.87\,cm across splits, indicating similar placement precision conditional on successful completion.

\begin{figure*}[t]
  \includegraphics[width=\textwidth]{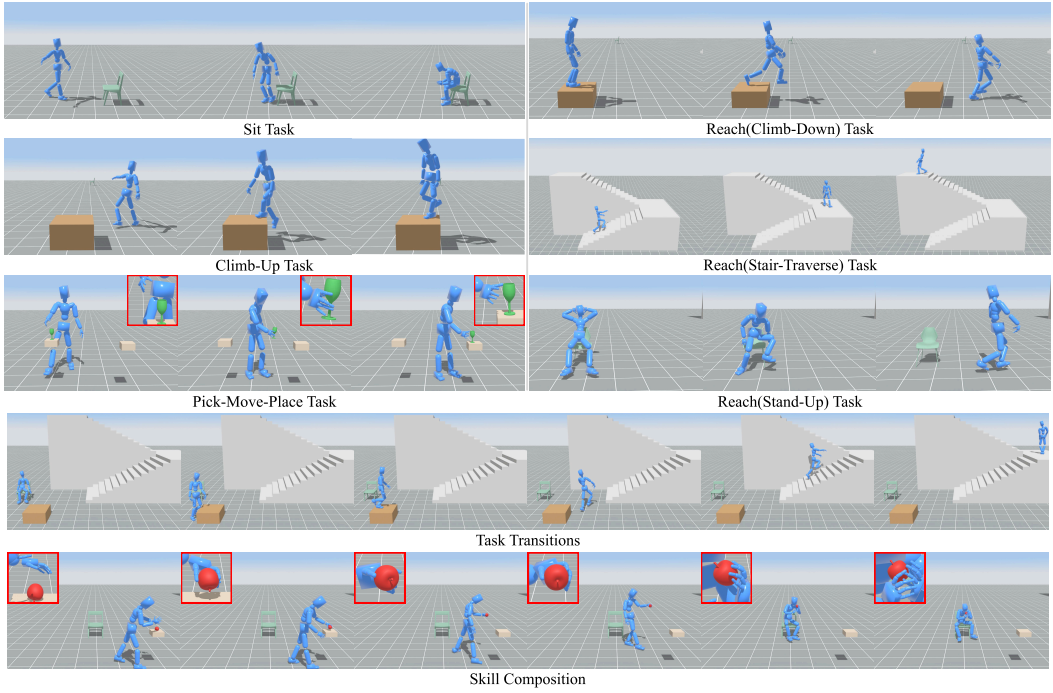}
  \vspace{-15pt}
  \caption{\textbf{Deployment of each task.} Our integrated policy demonstrates multiple tasks, encompassing scene-interaction to object manipulation. The model also covers transitioning between the tasks with compositing manipulation and scene-tasks.}
  \label{fig:application}
  \vspace{-5pt}
\end{figure*}

\subsection{Qualitative Results}
\label{subsec:qualitative-results}

Fig.~\ref{fig:application} qualitatively illustrates our policy performing multiple tasks in the environment containing both static structures and manipulable objects.
The character navigates toward designated goals, grasps and transports objects to target poses, and positions its body for interactions with the scene.
By specifying task goals and masks, we compose tasks learned during training into sequences of navigation, object manipulation, and body-scene interaction.
The illustrated rollouts exhibit smooth transitions between tasks while maintaining balance and coordinated full-body motion, consistent with the natural motion encouraged by the motion prior.

%% file: sec/07_conclusion.tex
\section{Conclusion}
\label{sec:conclusion}

In this work, we present a framework that unifies scene interaction and dexterous manipulation through a shared policy formulation.
A task-conditioned encoder predicts residual actions in a latent space defined by a frozen motion prior and decoder, while task and phase masks accommodate interaction-specific goals and observations switched at each defined task-phase.
Partial reference guidance and task-driven exploration support individual interactions, while cross-task training enables manipulation-to-scene transitions with continued grasp pose.

Our experiments demonstrate scene interaction and object manipulation under varied initial conditions, and qualitative results illustrate coordinated full-body motion across task transitions.
Among successful trials, placement precision remains comparable across evaluation splits.
These results show generalization to unseen object instances, while highlighting reliable interaction completion as a remaining challenge.

\paragraph{Future Work.}
Future work could incorporate visual and language-based task conditioning to support higher-level interaction specifications.
Extending the range of trained transitions and improving robustness to unseen objects could further support longer and more diverse task sequences.
Another interesting direction is to investigate how the learned motion prior and interaction policies can be adapted to real-world humanoids for the actual deployment.